\documentclass[sigconf]{acmart}

\usepackage{subcaption}   
\usepackage{placeins}
\usepackage{booktabs}
\usepackage{microtype}
\usepackage{xurl}
\usepackage{enumitem}

\setcopyright{none}
\renewcommand\footnotetextcopyrightpermission[1]{}
\usepackage{amsfonts} 
\newcommand{\StructSpace}{\mathbb{S}}
\usepackage{balance} 

\AtBeginDocument{%
  }

\newcommand{\model}{SRCO}

\title{Neural Structure Embedding for Symbolic Regression via Continuous Structure Search and Coefficient Optimization}

\author{Fateme Memar}
\affiliation{%
  \institution{Department of Electrical Engineering and Computer Science, University of Kansas}
  \country{USA}
}
\email{amemar@ku.edu}

\author{Tao Zhe}
\affiliation{%
  \institution{Department of Electrical Engineering and Computer Science, University of Kansas}
  \country{USA}
}
\email{taozhe@ku.edu}

\author{Dongjie Wang}
\affiliation{%
  \institution{Department of Electrical Engineering and Computer Science, University of Kansas}
  \country{USA}
}
\email{wangdongjie100@gmail.com}

\begin{document}

\begin{abstract}

Symbolic regression aims to discover human-interpretable equations that explain observational data. However, existing approaches rely heavily on discrete structure search (e.g., genetic programming), which often leads to high computational cost, unstable performance, and limited scalability to large equation spaces. To address these challenges, we propose SRCO, a unified embedding-driven framework for symbolic regression that transforms symbolic structures into a continuous, optimizable representation space. The framework consists of three key components: (1) structure embedding: we first generate a large pool of exploratory equations using traditional symbolic regression algorithms and train a Transformer model to compress symbolic structures into a continuous embedding space; (2) continuous structure search: the embedding space enables efficient exploration using gradient-based or sampling-based optimization, significantly reducing the cost of navigating the combinatorial structure space; and (3) coefficient optimization: for each discovered structure, we treat symbolic coefficients as learnable parameters and apply gradient optimization to obtain accurate numerical values. Experiments on synthetic and real-world datasets show that our approach consistently outperforms state-of-the-art methods in equation accuracy, robustness, and search efficiency. This work introduces a new paradigm for symbolic regression by bridging symbolic equation discovery with continuous embedding learning and optimization.

\end{abstract}

\maketitle
\pagestyle{plain}

\section{Introduction}
\label{sec:intro}

Many scientific and engineering domains require not only accurate
predictions but also explicit mathematical equations describing how input variables relate to an observed outcome.
Symbolic regression (SR) directly addresses this need by discovering human-interpretable
expressions from data without assuming a predefined model
form~\cite{koza1992gp,schmidt2009distilling,bongard2007reverseengineering,udrescu2020ai,la-cava2021srbench,srbench}.
These equations can be inspected, simplified, and compared with existing
theories, making SR particularly valuable in domains such as physics,
biology and chemistry, where understanding the underlying data-generating law is as important as predictive accuracy.

Despite its promise, symbolic regression remains challenging.
Classical SR approaches, including genetic programming and related evolutionary search methods,
formulate equation discovery as a combinatorial search problem over a discrete symbolic space~\cite{koza1992gp,ffx,operon2020gecco,virgolin2018gpgomea,randall2022bingo,la-cava2021srbench}.
This leads to two well-known difficulties.
First, small modifications to an expression tree can induce large and non-smooth changes in the resulting function, yielding unstable search
dynamics and poor scalability~\cite{koza1992gp,la-cava2021srbench}.
Second, the absence of a continuous notion of structural similarity limits reuse of previously discovered symbolic patterns,
forcing each new task to be solved largely from scratch. Recent neural and hybrid symbolic regression methods attempt to mitigate these issues
by introducing neural architectures to guide, generate, or prioritize symbolic expressions~\cite{petersen2019dsr,dso,nesymres,landajuela2022unified,martius2016eql,valipour2021symbolicgpt}.
While these approaches improve search heuristics or generation efficiency, they largely preserve the discrete formulation.
Expressions are still constructed token-by-token or tree-by-tree, and structural exploration remains entangled with coefficient fitting.
Consequently, challenges in stability, search efficiency, and structural generalization persist.
Existing symbolic regression approaches face three key limitations:

\begin{enumerate}
\item \textbf{Discrete symbolic representations lead to unstable and inefficient structure search.}
Operating directly on expression trees induces a highly non-smooth
optimization landscape, where local structural edits can produce
unpredictable functional changes, making principled optimization
difficult.

\item \textbf{Coupled optimization of structure and coefficients enlarges the effective search space.}
Many methods jointly search over operator choices, variable
selections, and numerical constants. This tight coupling increases
combinatorial complexity and can bias the search toward suboptimal
expressions due to conflicting optimization objectives.

\item \textbf{Lack of continuous structural representations limits reuse and guided exploration.}
Neural equation generators typically operate in token or tree spaces
and do not learn a continuous embedding capturing structural
similarity between symbolic expressions, limiting their ability to
generalize across datasets or difficulty regimes.
\end{enumerate}

\noindent\textbf{Our Approach.}
To overcome these limitations, we \emph{reformulate symbolic regression as a continuous structure optimization problem in a learned embedding space} (\textbf{SRCO}). SRCO decouples structure discovery from coefficient estimation and has three stages.

\begin{figure*}[htbp]
  \centering
  \makebox[\textwidth][c]{%
    \includegraphics[
      width=0.95\textwidth,
      keepaspectratio,
      trim=95 130 95 30,
      clip
    ]{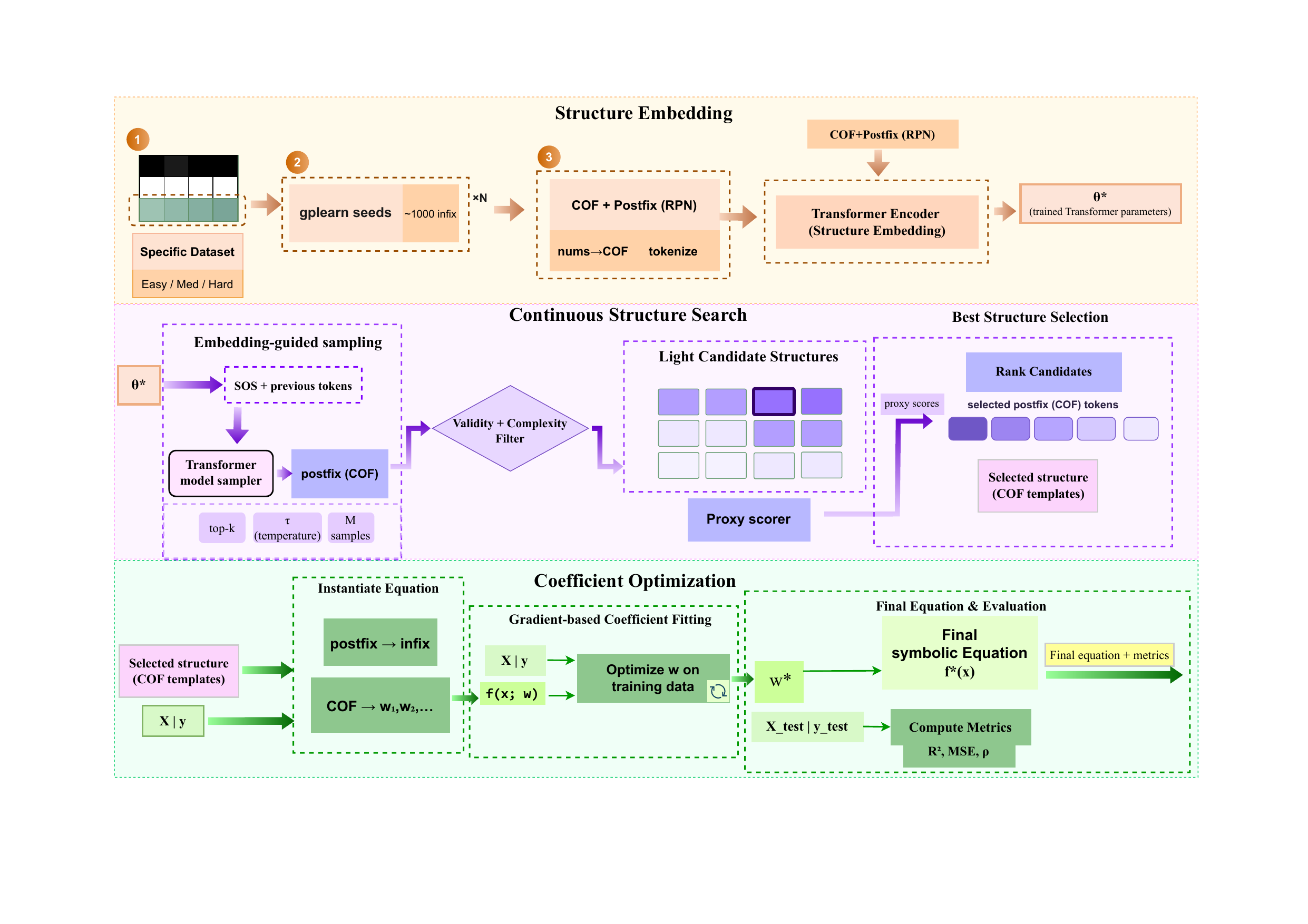}%
    }
\caption{Framework Overview of \textbf{SRCO}. \textit{Structure Embedding}: A GP-based SR system generates diverse candidate equations, which are converted into postfix sequences with abstracted coefficients (\texttt{COF}) to train a Transformer-based structural prior; \textit{Continuous Structure Search}: The learned prior guides constrained sampling in postfix space, followed by syntactic, semantic, and complexity filtering to obtain valid symbolic templates; \textit{Coefficient Optimization}: For each selected template, \texttt{COF} tokens are instantiated as learnable parameters and optimized via gradient-based regression to produce the final symbolic equation.}
\label{fig:pipeline}
\end{figure*}

\textbf{Structure Embedding:}
We first generate a large and diverse corpus of symbolic expressions
using a GP-based SR system~\cite{gplearn,koza1992gp}.
All numerical constants are abstracted into
a shared coefficient placeholder token (\texttt{COF}), and each
expression is converted to postfix (reverse Polish) notation.
A Transformer model is trained on these sequences to learn a
continuous embedding space that captures symbolic structure and defines
a probabilistic prior over valid equation templates~\cite{vaswani2017attention,lample2019deep,valipour2021symbolicgpt}.

\textbf{Continuous Structure Search:}
The learned structural embedding enables efficient exploration of the
symbolic space via constrained sampling.
New postfix expressions with \texttt{COF} placeholders are generated using
temperature-controlled decoding and top-$k$ sampling~\cite{noarov2025topk},
followed by syntactic, semantic, and complexity constraints.
This yields a pool of structurally valid equation templates without requiring discrete mutation or tree-based search.

\textbf{Coefficient Optimization:}
For each candidate structure, \texttt{COF} placeholders are
re-parameterized as learnable numeric coefficients and optimized on
the observed data via gradient-based or evolutionary optimization.
Final equations are selected based on predictive accuracy metrics under explicit complexity
constraints.
By decoupling structure discovery from coefficient optimization, the
proposed framework enables stable exploration in a continuous
representation space while leveraging efficient numerical fitting.
Our contributions can be summarized as follows:
\begin{itemize}[topsep=0pt,partopsep=0pt,itemsep=0pt,parsep=0pt,leftmargin=*,labelsep=0.5em]
\item \textbf{Problem.}
We reformulate symbolic regression as a continuous structure
optimization problem in a learned embedding space, exposing a
key bottleneck in existing discrete SR methods.
\item \textbf{Algorithm.}
We propose a modular framework that decouples symbolic structure
discovery from coefficient optimization, learning continuous
structure embeddings from postfix representations and enabling
embedding-guided exploration with gradient-based coefficient fitting.
\item \textbf{Evaluation.}
Experiments on synthetic and real-world benchmarks demonstrate
consistent improvements in accuracy, robustness, and search
efficiency over classical and neural symbolic regression methods,
while producing compact and interpretable equations~\cite{srbench,matsubara2024rethinking}.
\end{itemize}

\section{Related Work}
\label{sec:related-work}
Symbolic regression (SR) has a long history of discovering analytic expressions from data~\cite{koza1992gp,koza1994gp2,bongard2007reverseengineering,udrescu2020ai,la-cava2021srbench,srbench,banzhaf1998gpintro}.
Existing approaches differ primarily in how they represent symbolic structures, how they search over candidate equations, and how they optimize numerical coefficients. Below, we review these lines of work and highlight the representation-level limitations that motivate our framework.

\noindent\textbf{Discrete Structure Search in Symbolic Regression}
Classical symbolic regression methods formulate equation discovery as a combinatorial search problem over discrete symbolic structures, often expression trees.
Genetic programming (GP) and related evolutionary algorithms iteratively apply mutation and crossover operators to evolve a population of candidate expressions to improve
data fit~\cite{koza1992gp,schmidt2009distilling,la-cava2021srbench,operon2020gecco,virgolin2018gpgomea,randall2022bingo,miller2011cgp,ryan1998grammaticalevolution,hornby2006alps,deb2002nsga2}.
While these methods can explore rich hypothesis spaces and have achieved success in scientific discovery, their reliance on discrete representations leads to well-known challenges.
Small structural edits can induce large and unpredictable changes in the resulting function, yielding a highly non-smooth search landscape that resists principled optimization~\cite{koza1992gp,la-cava2021srbench,moraglio2012geometric,vanneschi2013gsgp}.
Moreover, structure and coefficients are often optimized jointly, forcing coefficient estimation to restart after structural changes and increasing search instability~\cite{vladislavleva2009order}.
A common mitigation is to combine discrete structure search with dedicated coefficient fitting or local solvers~\cite{ffx,pysr,keijzer2003improving,hansen2001cmaes}.
Although such approaches can improve numerical accuracy and data efficiency, the underlying search over symbolic structures remains discrete, and structure and coefficients remain tightly coupled.
As a result, these methods inherit many of the same scalability and stability limitations as classical GP-based SR.
\noindent\textbf{Neural Symbolic Regression and Token-Level Generation}
Recent work introduces neural models to guide or replace parts of the symbolic regression pipeline.
Methods such as Deep Symbolic Optimization (DSO), Neural Symbolic Regression that Scales (NeSymReS), and Transformer-based symbolic generation treat equations as token sequences and train neural sequence models to generate expressions auto-regressively~\cite{petersen2019dsr,dso,nesymres,vaswani2017attention,lample2019deep,valipour2021symbolicgpt}.
These models learn useful syntactic regularities and can leverage large synthetic corpora, and some approaches decouple constant fitting from sequence generation using dedicated numeric optimizers~\cite{dso,pysr,sahoo2018learning,trask2018nalu}.
 
Despite these advances, neural SR methods largely preserve the discrete nature of symbolic search.
Generation proceeds token-by-token, and structural exploration remains driven by discrete sampling rather than continuous optimization.
Crucially, these models typically do not provide a continuous embedding space explicitly optimized to capture structural similarity at the level of whole-equation templates, limiting their ability to reuse structural knowledge across datasets or to support guided exploration at the level of global equation structure.
\noindent\textbf{Representation Learning for Mathematical Structures}
In parallel, a body of work studies representation learning for mathematical objects such as expressions, programs, and logical formulas~\cite{lample2019deep,allamanis2018graphs,hellendoorn2020global,polu2020generative}.
These approaches encode symbolic structures as sequences, trees, or graphs and learn embeddings using recurrent networks, Transformers, or graph neural networks.
Learned representations have been successfully applied to tasks including formula manipulation and theorem proving~\cite{lample2019deep,polu2020generative}, and code modeling tasks such as completion and similarity~\cite{allamanis2018survey,hellendoorn2020global}.
However, prior representation-learning  approaches are designed for downstream predictive tasks rather than to drive an explicit symbolic regression pipeline.
Specifically, they do not integrate structural embeddings with (i) coefficient abstraction, (ii) a learned prior over equation templates, and (iii) a full loop for proposing, refining, and evaluating symbolic expressions on real observational data.
\noindent Our work bridges these lines by leveraging representation learning not as an auxiliary component, but as the foundation of symbolic regression itself.
By learning continuous embeddings of symbolic structure and using them to guide structure exploration, we decouple symbolic template discovery from coefficient optimization and enable a more stable and scalable approach to equation discovery.

\section{Problem Statement}
We study the problem of symbolic regression, where the goal is to recover an interpretable expression that explains the relationship between continuous inputs and a scalar output.
Given a dataset consisting of $N$ input--output pairs $\{(\mathbf{x}_i, y_i)\}_{i=1}^N$, where $\mathbf{x}_i \in \mathbb{R}^d$ and $y_i \in \mathbb{R}$.
In this paper, we represent a symbolic expression as the composition of a \emph{structure} and a set of \emph{numeric coefficients}.
The structure $S$ specifies the symbolic form of the expression and is represented in postfix notation with coefficient placeholder tokens (\texttt{COF}), while coefficients are instantiated as real-valued parameters $\mathbf{w} = (w_1, \dots, w_m)$, where $m$ is the number of \texttt{COF} placeholders in $S$.
Based on $S$ and $\mathbf{w}$, the resulting expression defines a function
$
  f_S(\mathbf{x}; \mathbf{w}) : \mathbb{R}^d \rightarrow \mathbb{R}.
$
This problem can be formulated as a nested optimization problem:
\[
  \min_{S \in \StructSpace} \; \min_{\mathbf{w}} \;
  \frac{1}{N} \sum_{i=1}^{N}
  \bigl(f_S(\mathbf{x}_i; \mathbf{w}) - y_i\bigr)^2
  \quad \text{s.t. } \mathcal{C}(S) \leq C_{\max}.
\]
Here, $\StructSpace$ denotes the space of admissible structures and
$\mathcal{C}(S)$ is a complexity measure (e.g., expression length or operator count),
with $C_{\max}$ specifying a user-defined complexity budget to ensure interpretability.

\section{Methodology}
\label{sec:method}

\subsection{Framework Overview}
Figure~\ref{fig:pipeline} demonstrates the framework overview of \model. In this paper, we propose a three-stage symbolic regression framework that explicitly decouples symbolic structure discovery from numeric coefficient optimization.
Given a dataset, the framework first learns a data-driven
structural prior over symbolic expressions by training a Transformer model on a large corpus of coefficient-abstracted postfix formulas~\cite{vaswani2017attention,lample2019deep,valipour2021symbolicgpt}.
This learned prior captures common structural patterns in valid
expressions and provides a continuous representation of symbolic structure.
In the second stage, the structural prior is used to guide the exploration of new symbolic structure templates via constrained sampling in symbolic space, producing a pool of candidate structures that satisfy syntactic, semantic, and complexity constraints.
In the final stage, numeric coefficients associated with each selected structure are instantiated and optimized using gradient-based regression on the original data.

\subsection{Structure Embedding}
\label{sec:step1-structure}

\subsubsection{Why structure embedding learning matters}
Symbolic regression requires reasoning over symbolic \emph{structures},
yet directly optimizing over the discrete space of expression trees or token sequences is inherently unstable: small structural changes can lead to large, non-smooth variations in the induced function, and the
search space lacks a meaningful notion of similarity between candidate
expressions~\cite{koza1992gp,la-cava2021srbench}.
Moreover, structural patterns discovered in one dataset
are difficult to reuse in others, as each search is performed largely
from scratch. These challenges point to a representation bottleneck:
effective symbolic regression requires a model that captures the
distribution of \emph{plausible symbolic structures} independently of
dataset-specific numeric coefficients.

To address this issue, we introduce a structure embedding stage that learns a continuous representation and a probabilistic prior over symbolic equation templates.
By abstracting away numeric constants and
modeling symbolic composition patterns directly, this stage captures shared structural regularities and provides a principled foundation for guided structure exploration in subsequent stages.

\subsubsection{Structural representation}
Let $S \in \StructSpace$ denote a symbolic structure, represented as a
postfix (reverse Polish) expression containing variable symbols,
operator/function symbols, and coefficient placeholders.
To disentangle
symbolic form from numeric values, we perform \emph{coefficient
abstraction}: all numeric constants in an expression are replaced by a
shared placeholder token \texttt{COF}. For example,
\[
  y = 3.2 \cdot x_0 + \sin(0.5 \cdot x_1)
  \;\Rightarrow\;
  y = \texttt{COF} \cdot x_0 + \sin(\texttt{COF} \cdot x_1).
\]
Each coefficient-abstracted expression is then deterministically
converted into postfix notation, yielding a token sequence
$\mathbf{t} = (t_1, \dots, t_L)$ with $t_\ell \in \mathcal{V}$, where
$\mathcal{V}$ denotes the vocabulary of variable symbols, operators,
\texttt{COF}, and special sequence markers. Malformed sequences are
discarded, and the length $L$ is bounded to enforce a complexity budget.

\subsubsection{Learning the structural prior}
To model the distribution of plausible symbolic structures, we first
construct a training corpus of equation templates by applying a
genetic-programming--based symbolic regression tool
(\texttt{gplearn})~\cite{gplearn} to each dataset. Using a fixed operator
set and a maximum expression depth, we generate candidate expressions
that achieve a minimal level of data fit. Aggregating candidates across
datasets yields a structurally diverse collection
$\{S_j\}_{j=1}^{K} \subset \StructSpace$, which serves as empirical samples
from the space of valid symbolic structures.

Each structure $S_j$ is represented as a postfix token sequence
$(t^{(j)}_1, \dots, t^{(j)}_{L_j})$. To learn a probabilistic model over
such sequences, we train a Transformer-based sequence model
parameterized by $\theta$~\cite{vaswani2017attention}. Each token $t^{(j)}_\ell$ is first mapped to a
learned embedding vector $\mathbf{e}^{(j)}_\ell \in \mathbb{R}^{d_{\text{model}}}$,
and positional encodings are added to preserve order information. The
embedded prefix sequence is then processed by the Transformer to
produce contextualized hidden representations,

\begin{equation}
  \bigl(\mathbf{h}^{(j)}_1,\dots,\mathbf{h}^{(j)}_\ell\bigr)
  =
  \mathrm{Transformer}_\theta
  \bigl(\mathbf{e}^{(j)}_1,\dots,\mathbf{e}^{(j)}_\ell\bigr).
\end{equation}

where $\mathbf{h}^{(j)}_\ell$ summarizes the symbolic context of the
expression prefix up to position $\ell$, and the model defines a
conditional distribution over the next symbolic token via a softmax
projection,
\begin{equation}
  p_\theta\bigl(
    t^{(j)}_{\ell+1} \mid t^{(j)}_1, \dots, t^{(j)}_\ell
  \bigr)
  =
  \mathrm{Softmax}\!\left( W \mathbf{h}^{(j)}_\ell + b \right),
\end{equation}
where $W$ and $b$ are learnable parameters. The Transformer is trained
autoregressively to maximize the log-likelihood of the observed postfix
sequences,

\begin{equation}
  \mathcal{L}_{\text{struct}}(\theta)
  =
  \sum_{j=1}^{K} \sum_{\ell=1}^{L_j-1}
  \log p_\theta\bigl(
    t^{(j)}_{\ell+1} \mid t^{(j)}_1, \dots, t^{(j)}_\ell
  \bigr).
\end{equation}

By conditioning each token on its preceding symbolic context, this
objective encourages the model to assign high probability to valid
symbolic compositions and low probability to invalid ones. As a result,
the learned distribution captures structural regularities of symbolic
expressions, including operator arity, permissible
operand--operator dependencies, and recurring expression motifs shared
across datasets. The trained sequence model defines a learned structural
prior $p_\theta(S)$ over $\StructSpace$ and induces a continuous embedding
space in which similar symbolic templates are mapped to nearby
representations. This structure prior is used to guide constrained
exploration of candidate structures in the next stages.

\subsection{Continuous Structure Search}
\label{sec:step2-search}

\subsubsection{Why prior-guided structure search matters}
Given a learned structural prior $p_\theta(S)$ over the symbolic
structure space $\StructSpace$, the goal of this stage is to turn
this prior into a practical search procedure that explores new symbolic
structures efficiently. Rather than relying on unguided random
mutations over expression trees, we use the prior to propose
structurally plausible candidates and steer selection using downstream
coefficient fitting and evaluation~\cite{koza1992gp,la-cava2021srbench}.

\subsubsection{Candidate proposal via prior sampling}
We treat the Transformer trained in the previous stage as a generative
model over symbolic structures represented in postfix notation with
coefficient placeholders \texttt{COF}. Structure generation proceeds
autoregressively: from a start-of-sequence token, the model
samples a token sequence

\[
\mathbf{t} = (t_1, \dots, t_T),
\]

until an end-of-sequence token is produced or $T=L_{\max}$ is reached. At each position $\ell$, the next token is sampled from the conditional distribution defined by the learned prior,

\begin{equation}
  t_\ell \sim p_\theta\bigl(t_\ell \mid t_1, \dots, t_{\ell-1}\bigr).
\end{equation}

We apply decoding controls including temperature scaling $\tau_{\text{temp}}$ and top-$k$ sampling~\cite{noarov2025topk},
and repeat the sampling process $M$ times to obtain a diverse set of candidate structures.
Each sampled token sequence induces a candidate symbolic structure $S \in \StructSpace$.

\subsubsection{Validity filtering and complexity control}
Not all sampled sequences correspond to valid or interpretable symbolic
expressions. Therefore, each candidate structure $S$ is passed through a
deterministic validation pipeline. First, a stack-based parser verifies
syntactic validity, ensuring that the postfix sequence reduces to
exactly one well-formed expression tree with correct operator arity.
Second, semantic constraints are enforced to discard expressions that
are guaranteed to be undefined over reasonable input ranges (e.g.,
trivial division-by-zero patterns). Finally, a complexity measure
$\mathcal{C}(S)$ is applied, and only structures satisfying
$\mathcal{C}(S) \leq C_{\max}$ are retained, where $C_{\max}$ is a predefined
interpretability budget.

For each sequence that passes validation, we obtain a symbolic structure
$S \in \StructSpace$ defined by its postfix representation. Each
occurrence of the coefficient placeholder \texttt{COF} is then replaced
by a distinct learnable parameter, yielding a parameterized symbolic
template
\begin{equation}
  f_S(\mathbf{x}; \mathbf{w}), \qquad
  \mathbf{w} = (w_1, \dots, w_m),
\end{equation}
where $m$ denotes the number of coefficient placeholders in $S$.
The resulting set of parameterized templates constitutes a pool of
candidate symbolic structures, which serves as the input to the
subsequent coefficient optimization stage.

\subsubsection{Proxy scorer: lightweight ranking during search}
To provide Stage~2 with a minimal-cost scoring signal, we attach a
lightweight \emph{proxy scorer} that assigns each valid candidate structure
a proxy score before coefficient optimization. In our implementation, this
proxy score is a decoding log-score accumulated during sampling: we sum the
log-probabilities of the sampled tokens under the temperature-scaled,
top-$k$ renormalized sampling distribution induced by $p_\theta$. This score
is inexpensive to compute and provides a structure-plausibility signal for
ranking candidates during search. Proxy scores are used only to rank/log
candidates; in all experiments we still optimize all validity/complexity-filtered
candidates unless stated otherwise. Final reporting is based on the post-fit
evaluation after coefficient optimization.

\subsubsection{Final scoring after coefficient optimization}
Candidate structures are then passed to coefficient optimization
(Section~\ref{sec:step3-coeff}). After fitting coefficients on the training
split, we compute standard regression metrics (MSE, $R^2$, and Pearson
$\rho$) and use these \emph{post-fit} scores to select the final equation
reported for each dataset.

\subsection{Coefficient Optimization}
\label{sec:step3-coeff}

\subsubsection{Why decoupled coefficient optimization matters}
Once a symbolic structure $S$ has been selected, the remaining task is
to determine numeric coefficients that best fit the observational data.
Optimizing coefficients with structure search is known to be
unstable and computationally inefficient, since structural changes invalidate prior parameter estimates~\cite{koza1992gp,la-cava2021srbench}.
By decoupling coefficient
optimization from structure discovery, we reduce the problem to a
standard continuous optimization task, enabling efficient
gradient-based methods while keeping the symbolic structure fixed.

\subsubsection{Instantiating coefficient placeholders}
Each validated symbolic structure $S$ produced in the previous stage
contains $m$ occurrences of the coefficient placeholder \texttt{COF}.
We replace these placeholders with distinct learnable parameters
$\mathbf{w} = (w_1, \dots, w_m)$, yielding a fully parameterized
expression
$
  f_S(\mathbf{x}; \mathbf{w}) : \mathbb{R}^d \rightarrow \mathbb{R}.
$
Under this formulation, the symbolic structure determines the functional
form of $f_S$, while $\mathbf{w}$ controls its numeric instantiation.

\subsubsection{Gradient-based coefficient fitting}
Given a dataset $\{(\mathbf{x}_i, y_i)\}_{i=1}^{N}$, we estimate
coefficients by minimizing the mean-squared error,
\begin{equation}
  \mathbf{w}^*
  =
  \arg\min_{\mathbf{w}}
  \mathcal{L}_{\mathrm{MSE}}(\mathbf{w})
  =
  \arg\min_{\mathbf{w}}
  \frac{1}{N} \sum_{i=1}^{N}
  \bigl(f_S(\mathbf{x}_i; \mathbf{w}) - y_i\bigr)^2.
\end{equation}

Because the symbolic structure is fixed, $f_S(\mathbf{x};\mathbf{w})$ is
a differentiable function of $\mathbf{w}$, allowing us to apply
gradient-based optimization (e.g., gradient descent). Optimization is
initialized from randomized coefficient values within a prescribed
range and is run with standard convergence safeguards, including a
maximum iteration budget and early stopping based on loss plateauing.

\subsubsection{Model selection and evaluation}
After coefficient optimization, each candidate equation is evaluated
using complementary performance metrics, including mean squared error
(MSE), coefficient of determination ($R^2$), and Pearson correlation
$\rho$. This evaluation is repeated for all candidate structures
produced by the structure search stage. For each dataset, we select a
final equation that achieves strong predictive performance under these
metrics while satisfying a predefined complexity constraint (e.g., maximum expression length or expression depth). By separating embedding-guided structure discovery from gradient-based coefficient optimization, the proposed framework helps isolate structural generalization from parameter estimation. This decoupled design improves optimization stability and interpretability, and enables fair comparison of coefficient refinement strategies.

\section{Experimental Results}
\label{sec:experiments}
\subsection{Experiment Setup}
\label{sec:exp-setup}

\subsubsection{Datasets}
\label{sec:exp-datasets}
We evaluate \textbf{SRCO} on two Feynman-based symbolic regression benchmarks~\cite{udrescu2020ai,la-cava2021srbench,srbench}, each organized into three difficulty tiers (easy/medium/hard). \textbf{Feynman--real-world} consists of real-world regression datasets derived from Feynman physics equations~\cite{udrescu2020ai}; we use the provided tiered subset with $30/40/50$ equations in the easy/medium/hard tiers and input dimensionalities $2/3/4$, respectively. For each equation, we use the official train/test split and the provided ground-truth expression; \textbf{SRCO} performs structure learning and coefficient refinement using only the training split, and all reported results are computed on the held-out test split. \textbf{Feynman--synthetic} is a synthetic variant constructed from the same equation collection~\cite{udrescu2020ai} following standard SR benchmark protocols~\cite{srbench}; for each equation (again $30/40/50$ in easy/medium/hard with $2/3/4$ inputs), we sample separate training and test sets directly from the ground-truth expression using the same function set as \textbf{SRCO}, i.e., $\{+,-,\times,\div,\sin,\cos\}$. \textbf{Structure corpus for training the prior.} For every equation in both benchmarks, we run a GP-based symbolic regression tool (\texttt{gplearn})~\cite{gplearn} on the training split to generate $M=1000$ candidate formulas. We use \texttt{SymbolicRegressor} with population size $2000$, $20$ generations, maximum depth $4$, parsimony coefficient $0.001$, and a fixed random seed, with the same primitive set $\{+,-,\times,\div,\sin,\cos\}$. After coefficient abstraction and postfix conversion (Section~\ref{sec:step1-structure}), the resulting templates form the structure corpus used to train the Transformer-based structural prior that drives \textbf{SRCO}'s subsequent structure search.

\subsubsection{Evaluation Metrics}
\label{sec:exp-metrics}
We evaluate each recovered equation $f(\mathbf{x})$ on the test split of its corresponding dataset using standard regression metrics.

\noindent
\textbf{(1) Mean Squared Error (MSE):}
\begin{equation}
  \mathrm{MSE}
  = \frac{1}{N_{\text{test}}}
    \sum_{i=1}^{N_{\text{test}}} (\hat{y}_i - y_i)^2,
\end{equation}
where $y_i$ is the true target and $\hat{y}_i = f(\mathbf{x}_i)$ is the predicted output.

\noindent
\textbf{(2) Coefficient of Determination ($R^2$):}
\begin{equation}
  R^2
  = 1 -
    \frac{\sum_{i=1}^{N_{\text{test}}} (\hat{y}_i - y_i)^2}
         {\sum_{i=1}^{N_{\text{test}}} (y_i - \bar{y})^2},
  \qquad
  \bar{y} = \frac{1}{N_{\text{test}}} \sum_{i=1}^{N_{\text{test}}} y_i,
\end{equation}
which measures the fraction of variance in the target explained by the recovered equation.

\noindent
\textbf{(3) Pearson Correlation ($\rho$):}
\begin{equation}
  \rho
  =
  \frac{\sum_{i=1}^{N_{\text{test}}} (\hat{y}_i - \bar{\hat{y}})
                                   (y_i - \bar{y})}
       {\sqrt{\sum_{i=1}^{N_{\text{test}}} (\hat{y}_i - \bar{\hat{y}})^2}
        \sqrt{\sum_{i=1}^{N_{\text{test}}} (y_i - \bar{y})^2}},
  \qquad
  \bar{\hat{y}} = \frac{1}{N_{\text{test}}} \sum_{i=1}^{N_{\text{test}}} \hat{y}_i,
\end{equation}
which captures the linear association between predictions and true outputs. For readability, when we report log-transformed errors (e.g., $\log(\mathrm{MSE})$), we always label the exact transform and its direction in the corresponding table/figure. \textbf{Per-equation evaluation and aggregation:} For each equation, \textbf{SRCO} samples candidate symbolic structures using the learned structural prior (Section~\ref{sec:step2-search}), fits coefficients on the training split (Section~\ref{sec:step3-coeff}), and evaluates the fitted expression on the test split. Unless otherwise noted, we report per-equation metrics and tier-averaged results for each benchmark.

\begin{table*}[t]
  \centering
  \caption{Overall performance on the Feynman--synthetic dataset.
  Average test performance over easy, medium, and hard tiers for SRCO and
  four baselines. Larger is better for $R^2$ and Pearson correlation $\rho$;
  smaller is better for $\log(\text{MSE})$.}
  \label{tab:feynman-synth}
  \begin{tabular}{lccc ccc ccc}
    \toprule
     & \multicolumn{3}{c}{Easy} & \multicolumn{3}{c}{Medium} & \multicolumn{3}{c}{Hard} \\
    \cmidrule(lr){2-4}\cmidrule(lr){5-7}\cmidrule(lr){8-10}
    Model &
    $R^2$ ($\uparrow$) & Pearson $\rho$ ($\uparrow$) & $\log(\text{MSE})$ ($\downarrow$) &
    $R^2$ ($\uparrow$) & Pearson $\rho$ ($\uparrow$) & $\log(\text{MSE})$ ($\downarrow$) &
    $R^2$ ($\uparrow$) & Pearson $\rho$ ($\uparrow$) & $\log(\text{MSE})$ ($\downarrow$) \\
    \midrule
    DSO & 0.58108 & 0.76836 & 45.0992 & -3.63397 & 0.57411 & 67.6205 & $-2.37841\times 10^{5}$ & 0.55307 & 67.0060 \\

    \texttt{gplearn}  & -0.25600 & 0.41807 & 45.0917 & 0.30200 & 0.42552 & 67.6205 & $-1.84\times 10^{13}$ & 0.39730 & 67.2514 \\
    FFX               & 0.45998 & 0.57570 & 45.0911 & 0.27657 & 0.61437 & 67.9408 & 0.26691 & 0.67575 & 68.4654 \\
    EFS               & -6.79227 & 0.06667 & 45.0915 & -21.75270 & 0.00000 & 67.6205 & -10.32860 & -0.12000 & 70.4763 \\
    \textbf{SRCO}     & 0.99821 & 0.96556 & 30.4281 & 0.99555 & 0.99786 & 67.14446 & 0.93708 & 0.94424 & 65.8075 \\
    \bottomrule
  \end{tabular}
\end{table*}

\begin{table*}[t]
  \centering
  \caption{Overall performance on the Feynman--real-world dataset.
  Average test performance over easy, medium, and hard tiers for SRCO and
  four baselines. Larger is better for $R^2$ and Pearson correlation $\rho$;
  smaller is better for $\log(\text{MSE})$.}
  \label{tab:feynman-real}
  \begin{tabular}{lccc ccc ccc}
    \toprule
     & \multicolumn{3}{c}{Easy} & \multicolumn{3}{c}{Medium} & \multicolumn{3}{c}{Hard} \\
    \cmidrule(lr){2-4}\cmidrule(lr){5-7}\cmidrule(lr){8-10}
    Model &
    $R^2$ ($\uparrow$) & Pearson $\rho$ ($\uparrow$) & $\log(\text{MSE})$ ($\downarrow$) &
    $R^2$ ($\uparrow$) & Pearson $\rho$ ($\uparrow$) & $\log(\text{MSE})$ ($\downarrow$) &
    $R^2$ ($\uparrow$) & Pearson $\rho$ ($\uparrow$) & $\log(\text{MSE})$ ($\downarrow$) \\
    \midrule
    DSO               & 0.59775 & 0.82824 & 45.0910 & -0.60127 & 0.59395 & 67.6205 & -6.17885 & 0.62223 & 66.9276 \\
    \texttt{gplearn}  & 0.13200 & 0.49617 & 45.0915 & -0.30200 & 0.44105 & 67.6205 & $-3.69 \times 10^{5}$ & 0.41335 & 67.0701 \\

    FFX               & 0.48157 & 0.61232 & 43.5123 & 0.25864 & 0.60612 & 67.9408 & 0.24350 & 0.64633 & 68.4654 \\
    EFS               & -0.64597 & 0.06667 & 45.0915 & -10.18850 & 0.00000 & 67.6205 & -38.91480 & -0.16000 & 70.4763 \\
    \textbf{SRCO}     & 0.95045 & 0.96984 & 39.7806 & 0.99936 & 0.99940 & 67.1605 & 0.94961 & 0.96152 & 65.8739 \\
    \bottomrule
  \end{tabular}
\end{table*}

\subsubsection{Baselines}
\label{sec:exp-baselines}
We compare \textbf{SRCO} against four established symbolic regression baselines. \textbf{Deep Symbolic Optimization (DSO)}~\cite{dso} is a reinforcement-learning SR method that constructs expressions token-by-token under a learned policy with a reward balancing fit and complexity. \textbf{Fast Function Extraction (FFX)}~\cite{ffx} is a deterministic sparse-regression approach that searches over a fixed library of candidate functions and selects a compact model. \textbf{Evolutionary Feature Synthesis (EFS)}~\cite{efs} is an evolutionary SR method that iteratively mutates and recombines candidate expressions under a complexity budget. Finally, \textbf{\texttt{gplearn}}~\cite{gplearn} serves as a GP baseline, where we report the best-fit expression returned directly by \texttt{gplearn} for each equation. All baselines are evaluated on the same Feynman--real-world and Feynman--synthetic benchmarks using identical train/test splits and the same test-set metrics (MSE, $R^2$, and Pearson $\rho$). Where baselines expose equivalent controls, we align the primitive operator set and enforce comparable complexity limits; otherwise, we use the closest available settings from the reference implementations and report the resulting performance under those configurations~\cite{srbench}.

\subsection{Experimental Results}
\label{sec:exp-results}

\subsubsection{Overall Performance}
\label{sec:overall-performance}

This experiment aims to answer the following question:
\emph{Does \textbf{SRCO} achieve better symbolic regression performance than existing SR baselines across both synthetic and real-world Feynman benchmarks?}
To answer this question, we evaluate \textbf{SRCO} and all baselines on Feynman--synthetic and Feynman--real-world and report tier-averaged test performance using $R^2$ ($\uparrow$), Pearson correlation $\rho$ ($\uparrow$), and $\log(\mathrm{MSE})$ ($\downarrow$). Tables~\ref{tab:feynman-synth} and~\ref{tab:feynman-real} summarize the overall results on Feynman--synthetic and Feynman--real-world, respectively~\cite{udrescu2020ai,srbench}.
We observe that \textbf{SRCO} consistently outperforms all baselines across easy, medium, and hard tiers on both benchmarks. On Feynman--synthetic (Table~\ref{tab:feynman-synth}), \textbf{SRCO} achieves strong fits across all tiers (e.g., $R^2 \approx 0.998/0.996/0.94$ from easy to hard) and attains the lowest $\log(\mathrm{MSE})$ in every tier. In contrast, DSO and FFX degrade substantially as difficulty increases (including negative $R^2$ on medium/hard for DSO), while EFS and \texttt{gplearn} are consistently worse. A similar trend holds on Feynman--real-world (Table~\ref{tab:feynman-real}): \textbf{SRCO} achieves the best test performance across tiers, with high correlation and favorable error values, while the baselines again lag behind and often deteriorate on the more difficult tiers. We attribute these gains to two key aspects of \textbf{SRCO}: (i) structure learning from a large corpus of coefficient-abstracted postfix templates, which helps the model reuse effective structural motifs across equations; and (ii) decoupled coefficient refinement on the training split, which improves numerical fit without destabilizing the learned structures. In summary, Tables~\ref{tab:feynman-synth} and~\ref{tab:feynman-real} demonstrate that \textbf{SRCO} delivers the most reliable performance across both benchmarks while remaining competitive in expression complexity.

\begin{figure}[t]
  \centering
  \includegraphics[width=\columnwidth]{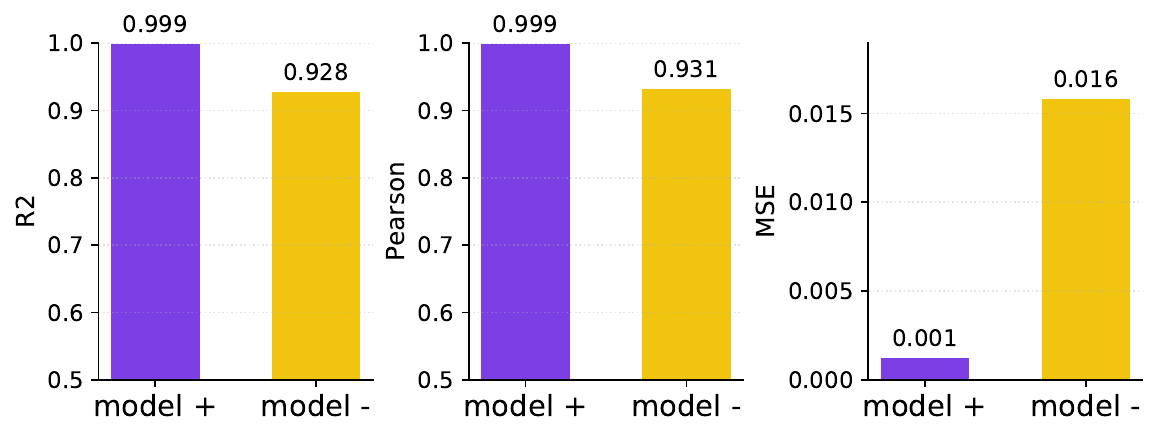}
\caption{Ablation of coefficient optimization on Feynman-bonus.1. We compare SRCO’s gradient-based coefficient fitting (\texttt{model+}) to stochastic hill-climbing (random search; \texttt{model-}) while keeping the template, train/test split, and optimization budget fixed. Bars report held-out test-set performance (higher is better for $R^2$ and $\rho$, lower is better for MSE).}

  \label{fig:coef-ablation}
\end{figure}

\subsubsection{Ablation Study: Coefficient Optimization}
\label{sec:ablation-study}

This experiment aims to answer the following question:
\emph{Is gradient-based coefficient fitting necessary for \textbf{SRCO} to achieve high-fidelity symbolic regression?}
To answer this question, we construct an ablation variant of \textbf{SRCO} that replaces our gradient-based coefficient fitting with a stochastic hill-climbing (random search) procedure, while keeping all other components fixed (the sampled structure template, the training/test split, and the optimization budget). Both optimizers fit coefficients on the \textbf{training split} of the same equation instance (Feynman-bonus.1) under an identical budget, and we report performance on the held-out \textbf{test split} in Fig.~\ref{fig:coef-ablation}. We observe that gradient-based fitting achieves a near-perfect fit with $R^2 = 0.999$, Pearson correlation $\rho = 0.999$, and $\mathrm{MSE} \approx 0.001$, whereas hill-climbing saturates at $R^2 = 0.928$, $\rho = 0.931$, and $\mathrm{MSE} \approx 0.016$. A plausible reason is that gradient-based updates use local curvature information to refine continuous coefficients efficiently, while hill-climbing relies on random perturbations and accepts a candidate only when it reduces the \textbf{training loss}, which can be less sample-efficient in higher-dimensional coefficient spaces. In summary, differentiable coefficient optimization is a key contributor to \textbf{SRCO}'s final accuracy under the same budget, yielding substantially stronger fits even when the symbolic structure is held constant.

\subsubsection{Robustness to Noise}
\label{sec:robustness-noise}

This experiment aims to answer: \emph{How robust is \textbf{SRCO} to feature corruption at test time?}
To answer this question, we inject additive Gaussian noise into the \emph{test} input features while keeping the targets $y$ and the test split fixed~\cite{bishop1995training}. Specifically, for noise level $\eta \in [0,1]$, we evaluate on perturbed inputs $\tilde{\mathbf{x}}=\mathbf{x}+\boldsymbol{\epsilon}$ where $\epsilon_{ij}\sim\mathcal{N}(0,(\eta\sigma_j)^2)$ and $\sigma_j$ is the empirical standard deviation of feature $j$ \emph{computed on the training split}. We keep the discovered equation (structure and fitted coefficients) fixed and only perturb the test inputs at evaluation time. Table~\ref{tab:noise-robustness} reports held-out test performance across noise levels from $0\%$ to $100\%$ using $R^2$, Pearson correlation $\rho$, and $\mathrm{MSE}$; we report $\log(\mathrm{MSE})$ in the table for readability. We observe that performance degrades smoothly as noise increases: $R^2$ remains $\ge 0.99$ up to $10\%$ noise, stays $\ge 0.98$ through $40\%$, and remains $\ge 0.95$ through $70\%$, while correlation stays high even under substantial corruption. Consistently, $\log(\mathrm{MSE})$ increases monotonically with noise level (i.e., becomes less negative), indicating predictable error growth without instability. In summary, \textbf{SRCO} retains near-perfect fit under mild noise and continues to produce meaningfully predictive outputs even under severe feature corruption.

\begin{table}[t]
\centering
\small
\setlength{\tabcolsep}{5pt}
\begin{tabular}{@{}rccc@{}}
\toprule
\textbf{Noise (\%)} &
\textbf{$\log(\mathrm{MSE})$ $\downarrow$} &
\textbf{$R^2$ $\uparrow$} &
\textbf{Pearson $\rho$ $\uparrow$} \\
\midrule
0   & -24.5128 & 0.9990 & 0.9990 \\
10  & -23.6677 & 0.9973 & 0.9976 \\
20  & -23.2118 & 0.9941 & 0.9995 \\
30  & -22.8896 & 0.9893 & 0.9916 \\
40  & -21.6436 & 0.9828 & 0.9868 \\
50  & -20.4409 & 0.9745 & 0.9802 \\
60  & -19.2661 & 0.9643 & 0.9718 \\
70  & -18.1114 & 0.9521 & 0.9610 \\
80  & -16.9725 & 0.9376 & 0.9472 \\
90  & -14.8458 & 0.9206 & 0.9300 \\
100 & -12.7296 & 0.9009 & 0.9009 \\
\bottomrule
\end{tabular}
\caption{Robustness to feature noise. We add Gaussian noise to the \emph{test} input features (scaled by per-feature standard deviations computed on the training split) and evaluate a fixed equation on the held-out test set. We report $\log(\mathrm{MSE})$ for readability, along with $R^2$ and Pearson correlation $\rho$.}
\label{tab:noise-robustness}
\end{table}

\subsubsection{Inference Efficiency}
\label{sec:inference-efficiency}

\begin{figure}[t]
  \centering
  \includegraphics[width=\columnwidth]{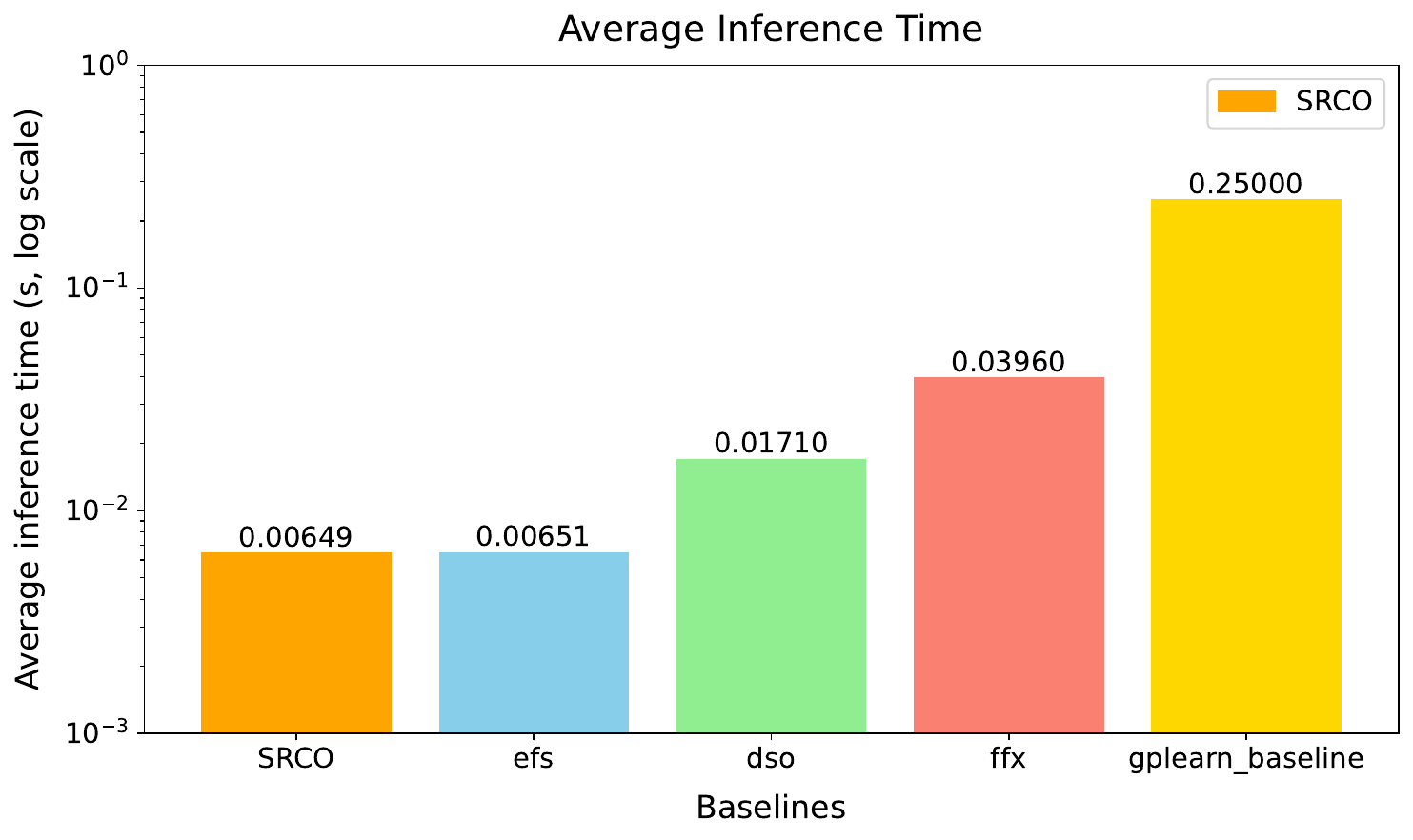}
  \caption{Average per-equation equation-evaluation time on the \textbf{test split} (seconds; lower is better), averaged over six settings (2 benchmarks $\times$ 3 tiers: Feynman--synthetic/real-world $\times$ easy/medium/hard). \textbf{SRCO} achieves the fastest evaluation (0.00649\,s), essentially tied with EFS (0.00651\,s) and outperforming DSO (2.6$\times$ slower), FFX (6.1$\times$), and \texttt{gplearn} (38.5$\times$), while maintaining strong accuracy (Tables~\ref{tab:feynman-synth}--\ref{tab:feynman-real}).}
 
  \label{fig:inference-time}
\end{figure}

This experiment aims to answer: \emph{How efficient is \textbf{SRCO} at test-time equation evaluation compared with SR baselines?}
To answer this question, we measure per-equation \emph{inference} time as a single forward evaluation of each method’s final discovered equation on the \textbf{test split} (no retraining). We report wall-clock time averaged over six settings (2 benchmarks $\times$ 3 tiers: Feynman--synthetic/real-world $\times$ easy/medium/hard), using one representative equation instance per setting. Figure~\ref{fig:inference-time} visualizes the resulting averages. \textbf{SRCO} achieves the fastest evaluation time at $0.00649$\,s per equation, essentially matching EFS ($0.00651$\,s) and substantially outperforming DSO ($0.0171$\,s; 2.6$\times$ slower), FFX ($0.0396$\,s; 6.1$\times$), and the \texttt{gplearn} baseline ($0.250$\,s; 38.5$\times$). A likely reason is that \textbf{SRCO} produces compact closed-form expressions with minimal per-sample overhead, whereas several baselines incur heavier symbolic evaluation costs. In summary, \textbf{SRCO} offers a favorable accuracy--efficiency trade-off, maintaining high test performance (Tables~\ref{tab:feynman-synth}--\ref{tab:feynman-real}) while keeping evaluation cost minimal.

\subsubsection{Parameter Sensitivity}
\label{sec:parameter-sensitivity}

\begin{figure}[!t]
  \centering
  \includegraphics[width=0.88\linewidth]{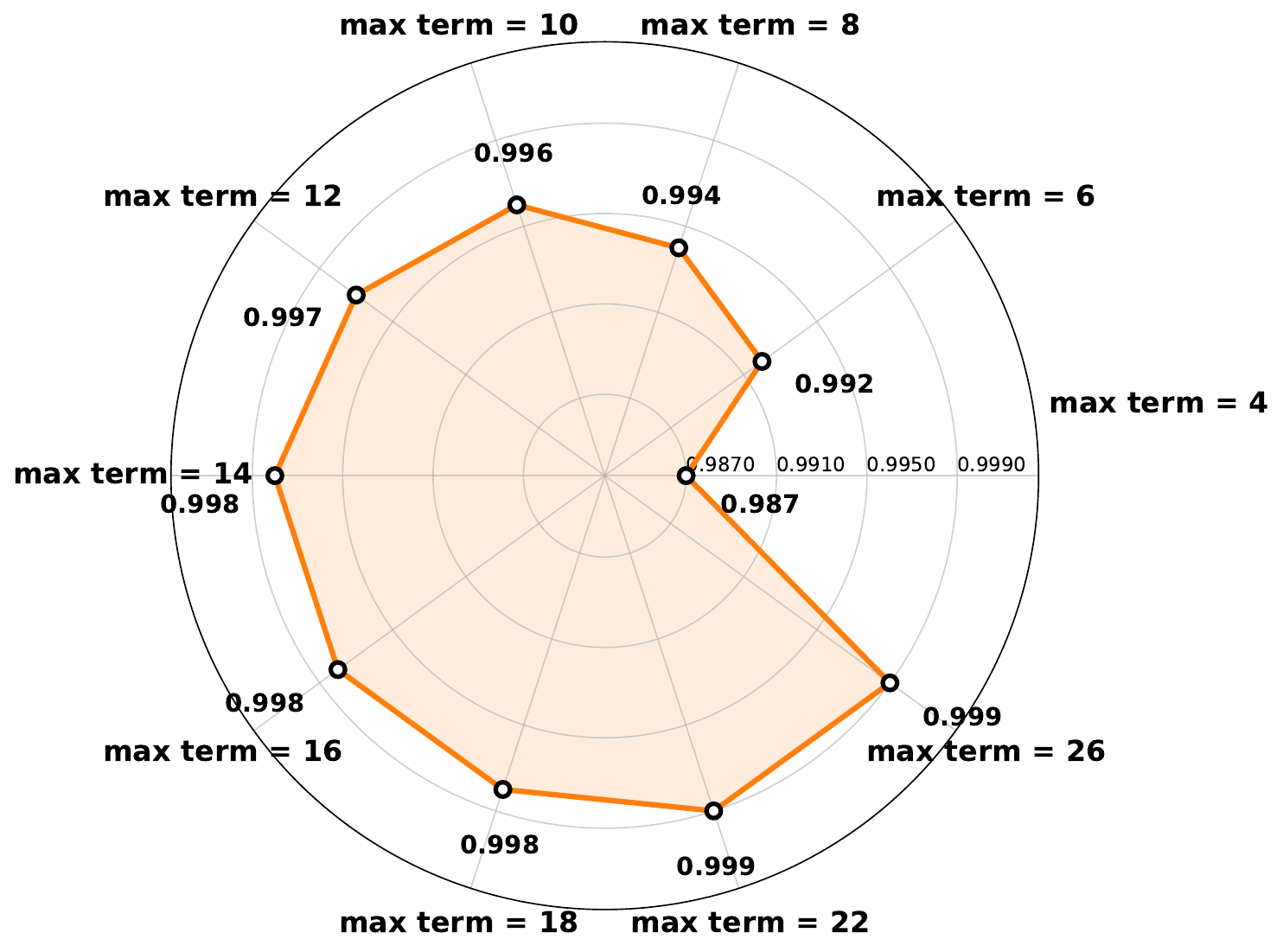}
  \caption{\small Pearson correlation $\rho$ for \texttt{max\_term}. Accuracy improves monotonically and saturates around 18--22 terms.}
  \label{fig:maxterm-pearson}
\end{figure}

\begin{figure}[!t]
  \centering
  \includegraphics[width=0.88\linewidth]{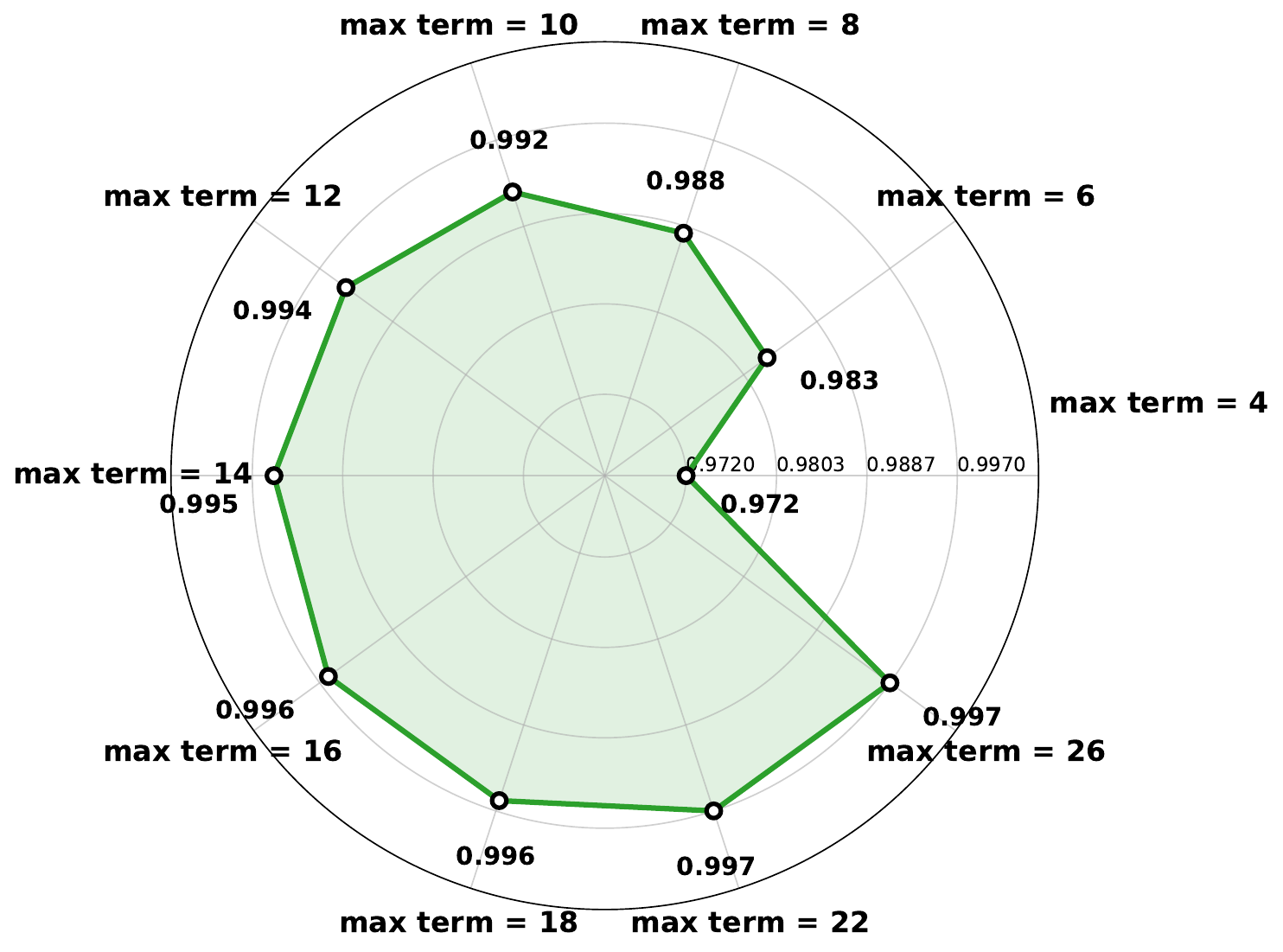}
  \caption{\small $R^2$ for \texttt{max\_term}. Results mirror Pearson correlation $\rho$, with diminishing returns after 18--22 terms.}
  \label{fig:maxterm-r2}
\end{figure}

\begin{figure}[!t]
  \centering
  \includegraphics[width=0.88\linewidth]{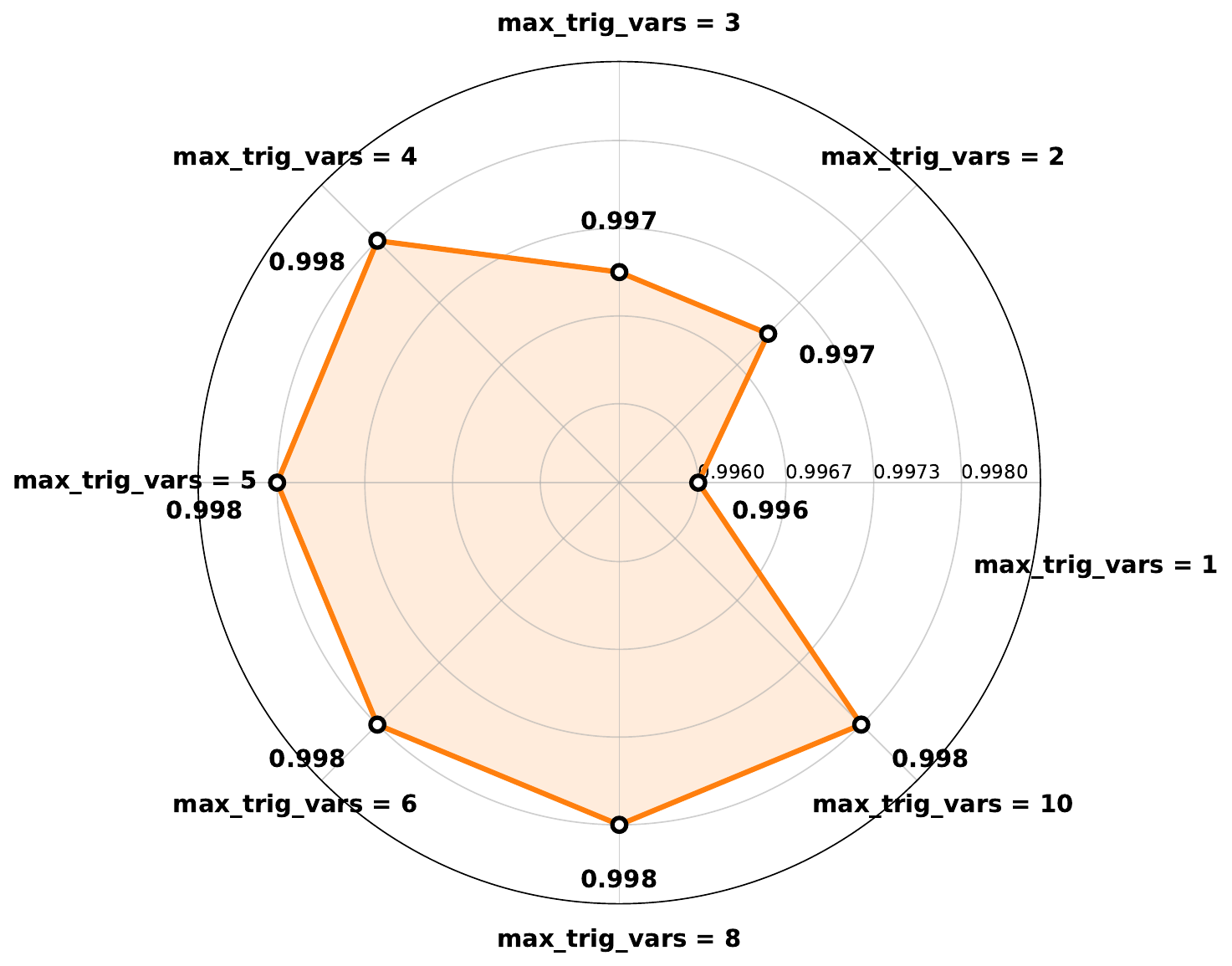}
  \caption{\small Pearson correlation $\rho$ for \texttt{max\_trig\_vars}. Most gains accrue by about 3--4 trig variables; larger budgets provide marginal change.}
  \label{fig:maxtrig-pearson}
\end{figure}

\begin{figure}[!t]
  \centering
  \includegraphics[width=0.88\linewidth]{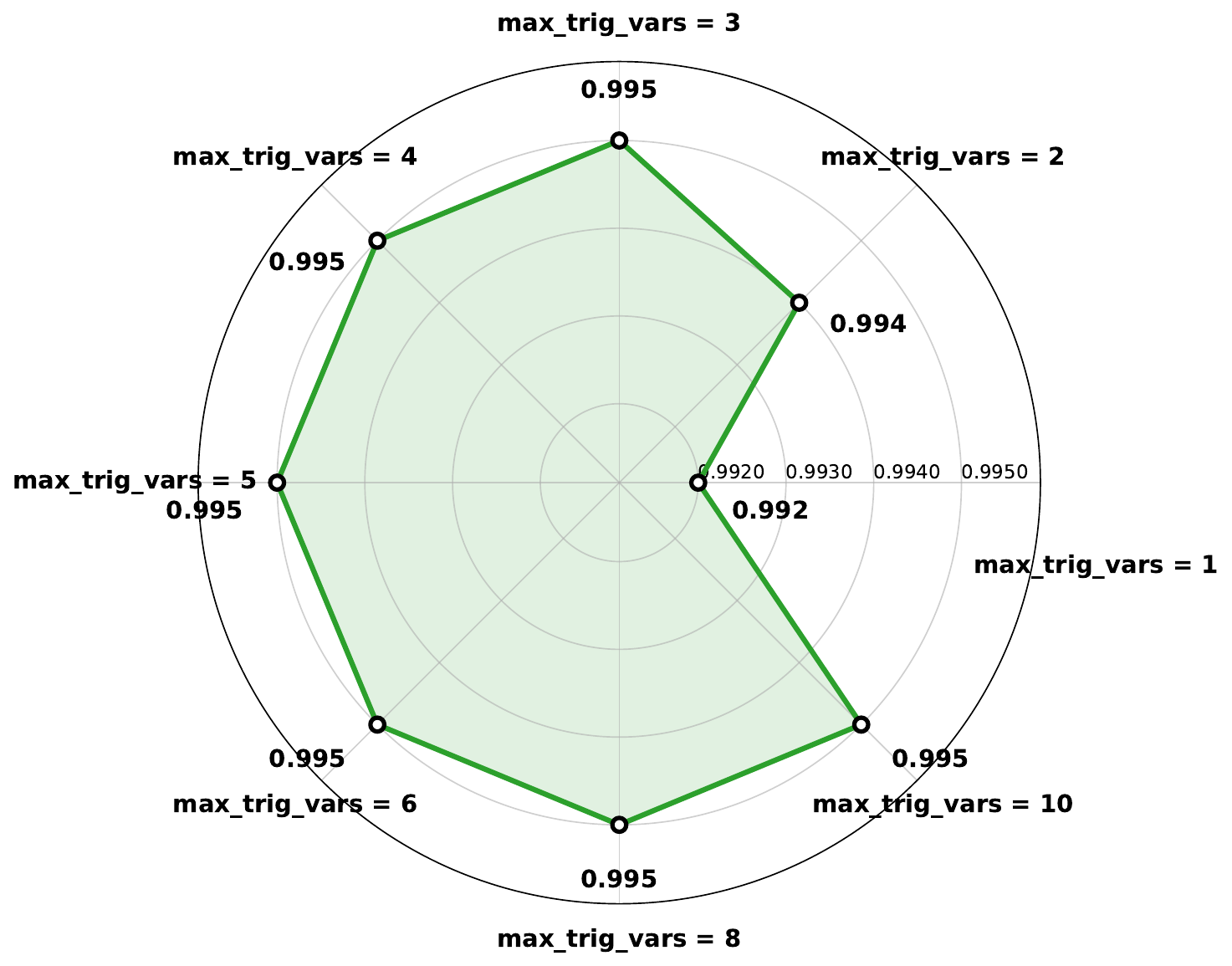}
  \caption{\small $R^2$ for \texttt{max\_trig\_vars}. Performance improves quickly at small budgets and largely saturates by about 3--4, with only incremental differences through 6--10.}
  \label{fig:maxtrig-r2}
\end{figure}

This experiment aims to answer: \emph{How sensitive is \textbf{SRCO} to key structure-search hyperparameters that control template expressiveness?}
To answer this question, we vary one hyperparameter at a time keeping the remaining components of \textbf{SRCO} fixed (structure prior, sampling constraints, and coefficient optimization), and evaluate each configuration on the same train/test split using test-set $R^2$ and Pearson correlation $\rho$. Results are summarized in the radar plots in Figures~\ref{fig:maxterm-pearson}--\ref{fig:maxterm-r2} and Figures~\ref{fig:maxtrig-pearson}--\ref{fig:maxtrig-r2}. Increasing \texttt{max\_term} (the maximum number of terms allowed in the sampled expression template) consistently improves performance: Pearson correlation rises from $0.987$ at \texttt{max\_term}$=4$ to $0.999$ at \texttt{max\_term}$=26$, and $R^2$ increases from $0.972$ to $0.997$, with diminishing returns beyond roughly 18--22 terms (Figures~\ref{fig:maxterm-pearson},~\ref{fig:maxterm-r2}). For \texttt{max\_trig\_vars} (the maximum number of trigonometric variables/operators allowed in the template), performance improves quickly at small budgets and then saturates: the main gains occur by about 3--4 trig variables, while larger settings (e.g., 6--10) yield only marginal changes (Figures~\ref{fig:maxtrig-pearson},~\ref{fig:maxtrig-r2}). The underlying driver is that increasing these budgets expands the candidate template space, but once sufficient capacity is available to express the target structure, additional capacity provides limited benefit. In summary, \textbf{SRCO} is stable across a wide range of values, and practical near-ceiling settings are \texttt{max\_term} $\approx$ 18--22 and \texttt{max\_trig\_vars} $\approx$ 3--6.

\balance
\section{Conclusion}
\label{sec:conclusion}

We presented \textbf{SRCO}, a three-stage symbolic regression framework that decouples structure discovery from coefficient optimization.
 Given observational datasets of varying difficulty, we use a GP-based SR tool to generate candidate equations, abstract numeric constants into a coefficient token (COF), and convert expressions into postfix sequences; a Transformer learns a continuous structural embedding and a prior over valid templates. We then use this prior to explore candidate structures via constrained sampling and a validation pipeline that enforces syntactic correctness and semantic and complexity constraints. Validated templates are reparameterized so each COF becomes an independent coefficient, and coefficients are fit with gradient-based optimization on the training split; final equations are selected under complexity budgets using MSE, $R^2$, and Pearson correlation. Across easy, medium, and hard tiers on both Feynman--synthetic and Feynman--real-world benchmarks, SRCO achieves strong predictive accuracy while maintaining interpretability and competitive inference efficiency relative to established SR baselines. Combining a Transformer-guided structural prior with a separate gradient-based coefficient stage is an effective and scalable design pattern for symbolic regression.
Our evaluation is limited to two Feynman benchmarks; next steps include expanding experiments to broader benchmark suites and higher-dimensional settings with richer operator libraries, reducing reliance on \texttt{gplearn}-bootstrapped template corpora for template construction, and extending structure search beyond constrained sampling through complementary proposals and multi-objective selection under fixed computational budgets.



\FloatBarrier
\balance
\bibliographystyle{ACM-Reference-Format}
\bibliography{sample-base}

\end{document}